 \documentclass[final,1p,times]{elsarticle}

\usepackage{graphicx}
\usepackage{amssymb}
\usepackage{supertabular}
\usepackage{amsmath}
\usepackage{algorithmic}
\usepackage{tabularx}
\usepackage{multicol}
\usepackage{lineno}
\usepackage{booktabs}
\usepackage{lscape}

\journal{Information Sciences}

\begin{document}

\begin{frontmatter}

\title{Cuckoo Search Inspired Hybridization of the Nelder-Mead Simplex Algorithm Applied to Optimization of Photovoltaic Cells}

\author[add1,add2]{Raka Jovanovic}
\address[add1]{Qatar Environment and Energy Research Institute (QEERI)}
\address[add2]{Institute of Physics, University of Belgrade, Pregrevica 118, Zemun, Serbia}
\ead{rjovanovic@qf.org.qa}

\author[add1,add3]{Sabre Kais}
\address[add3]{Department of Chemistry,  Physics and Birck Nanotechnology Center, Purdue University, West Lafayette, IN 47907 US}

\author[add1,add4]{Fahhad H Alharbi}
\address[add4]{King Abdulaziz City for Science and Technology (KACST)}

\begin{abstract}
A new hybridization of the Cuckoo Search (CS) is developed and applied to optimize multi-cell solar systems; namely multi-junction and split spectrum cells. The new approach consists of combining the CS with the Nelder-Mead method. More precisely, instead of using single solutions as nests for the CS, we use the concept of a simplex which is used in the Nelder-Mead algorithm. This makes it possible to use the flip operation introduces in the Nelder-Mead algorithm instead of the Levy flight which is a standard part of the CS. In this way, the hybridized algorithm becomes more robust and less sensitive to parameter tuning which exists in CS. The goal of our work was to optimize the performance of multi-cell solar systems. Although the underlying problem consists of the minimization of a function of a relatively small number of parameters, the difficulty comes from the fact that the evaluation of the function is complex and only a small number of evaluations is possible. In our test, we show that the new method has a better performance when compared to similar but more compex hybridizations of Nelder-Mead algorithm using genetic algorithms or particle swarm optimization on standard benchmark functions. Finally, we show that the new method outperforms some standard meta-heuristics for the problem of interest.
\end{abstract}





\end{frontmatter}

\section{Introduction}
In the recent years, the performance of photovoltaic (PV) cells has been significantly improved by using multi-cell devices. In such systems, few different cells are combined to maximize the conversion efficiency by dividing the spectrum of solar radiation and then use a proper cell for each divided spectrum. Using this approach, efficiency of 44.7\% has been achieved where 4 junctions are used \cite{ADDED01,ADDED02}. The two most common methods for such device concept are splits spectrum and multi-junction solar cells. With the development of new technologies, it has become possible to create more complex systems consisting of a higher number of cells. It has  been shown that  the performance of such systems can be further improved using optical concentration\cite{FahhadSSEF}. For future development of such solar-cell systems, it is of significant importance to have bounds for the optimal possible efficiency. While it is relatively simple to find them in case of two or three layers, it becomes significantly more complex in case of a higher number of layers. This is due to the fact that it is necessary to find the minimal value of a multi parameter function, which is computationally challenging.

The problem becomes even more complex because it is hard to find the corresponding gradient that could simplify the calculation. There is a wide range of non-gradient based optimization methods like simulated annealing \cite{Kirkpatrick13051983}, genetic algorithms \cite{ADDED03}, pivot method \cite{serra1997pivot}, particle swarm optimization \cite{PSO}, Nelder-Mead Simplex (NMS) method \cite{Nelder01011965}  that are generally used to solve this type of problems. The performance of such methods is highly dependent on the function that we wish to minimize. In the case of the problem of interest, initial test have shown that NMS algorithm manages to outperform, in several tested problem instances, mentioned more complex population based methods. One of the reasons for this is the fact that due to physical properties of the problem, we have a good initial guess of the solution. Our research has focused on improving the performance of this algorithm by incorporating some type of swarm intelligence. There is a wide range of meta heuristic that mimic the behavior of groups in nature, the most prominent are ant colony optimization \cite{Dorigo:2004:ACO:975277}, bee colony optimization \cite{BEESR}, particle swarm optimization (PSO)  \cite{PSO}, firefly algorithm \cite{Subotic:2012:OIF:2209654.2209732}, and cuckoo search (CS)\cite{yang2009cuckoo}\cite{yang2010engineering}.  In the recent years, the CS \cite{yang2009cuckoo} algorithm has been gaining on popularity as an optimization method due to its good performance, robustness, and simplicity of implementation. Research has been conducted to improve the original algorithm by changing the method of communication inside of the colony \cite{Rajabioun20115508}, having adaptive parameters\cite{Walton2011710}, using quantum inspired approaches\cite{layeb2012novel}\cite{layeb2011novel}, and adapting the algorithm for parallel applications \cite{Subotic:2012:PCS:2230596.2230624}\cite{jovanovicparallelization}. The CS algorithm has previously been successfully applied in the field of solar cells, more precisely for parameter estimation of photovoltaic models \cite{CSPhoto}. 

Previously, the NMS algorithm has been hybridized using genetic algorithms \cite{NMSGA}\cite{Chelouah2003335}, ant colony optimization \cite{NMSACO} and particle swarm optimization \cite{Zahara20093880} resulting in very efficient global optimization algorithms.  One of the main problems of hybridized methods is that although they achieve better results, they often become very complex for implementation. In this work, we introduce a CS inspired hybridization of the NMS algorithm that manages to avoid this drawback but still achieves significantly better results than the original method. The general idea of our approach is to use the simplex structure from NMS algorithm as a form of a nest in the cuckoo search. One of the main advantages of this approach is that we can avoid the use of the Levy flight in CS. Although the Levy flight is a very powerful system of exploring the solution space, its performance within the CS is highly dependent on the scaling factor introduced in this method. In the presented approach, we substitute the Levy flight with the flip operation of a simplex in the in the NMS algorithm. In our tests, we show that the proposed method not only performs well in the case of solar cells optimization, but also achieves good results on standard benchmark functions when a lower number of parameters is considered. The cuckoo search is often a competing method to the particle swarm optimization and genetic algorithms (GA); because of this, we also give a comparison to previously published results of hybridization of NMS using this method. In  the comparison we have shown that the new method manages to achieve a higher speed of convergence, while at the same time being much simpler to implement. To better evaluate the effect of the proposed hybridization of CS we have also conducted a comparison to other versions of the CS previously presented in literature. This has been done on a wide range of benchmark functions, from which it is noticeable that the new approach has a very good performance.

The  article is organized as follows. In the next section, we give an overview of related research,  divided into 3 subsections. The first one presents the basics of the cuckoo search algorithm. In the second subsection we give an outline of the NMS algorithm.  In the final subsection we give an overview of previously developed hybridization of NMS using swarm intelligence and genetic algorithms.   The third section presents a detailed specification of the new method. The forth section is dedicated to an evaluation of the new method. In the first subsection we compare its performance on standard benchmark functions with published results for other population based hybridizations of NMS. The following subsection presents a comparison to other versions of the CS algorithm. In the third subsection we present the optimization problem in PV cell and  analyze the performance of the method.
\section{Related work}
\subsection{Cuckoo Search}

As previously mentioned the Cuckoo Search is one of the population based meatheuristic for function minimization. In other words the goal of this method is to find the global minimal value  $min f(x)$ of a function defined in the following way 

\begin{equation}
\label{Deff}
f: \mathbb{R}^n \rightarrow \mathbb{R}
\end{equation} 

Function $f$ is called the objective function with $n$ dimensions.

Cuckoo search (CS) is an optimization algorithm that mimics the brood parasitism of some cuckoo species.  More precisely, cuckoo birds lay their eggs in the nests of other host birds (of other species). Through an evolutive process cuckoos have managed to adapt the shape and color of eggs to the one of targeted bird species.  This method of survival has been converted to a meta-heuristic approach called Cuckoo Search. In the corresponding algorithm, each egg in a nest will represent a solution and cuckoo egg represents a newly generated solutions. The idea is to create new, similar, and potentially higher quality solutions (cuckoos) to replace the low quality solutions in the nests. In the simplest form, each nest contains one egg.

The proposed meta heuristic follows several rules.
\begin{enumerate}
\item Each cuckoo in the colony generates one solution (egg) at each step, similar to an already existing solution (nest), \smallskip

\item The best solution will be carried to the next generation, \smallskip

\item The number of available solutions that are used as nests is fixed, and at each step, new solutions are generated. This has a consequence that some solutions need to be discarding, which corresponds to eggs being found by the host bird.  The discarding operation is only done on some set of worst solutions (nests), more precisely for each such nest $a$ there is a probability $p_a$ of being found.
\end{enumerate}

The standard CS algorithm following these rules is presented in the following pseudo code:

\begin{algorithmic}
\STATE
\STATE{Objective function: $f(X)$, $X=(x_1, x_2,.., x_d)$}
\STATE
\STATE Generate an initial population of  $n$ host nests;
\WHILE{($t<MaxGeneration$) or ($stop criterion$)}
   \STATE Get a cuckoo randomly (say, $i$) and replace its solution by performing Levy flights;
   \STATE Evaluate its quality/fitness $F_i$
    \STATE
   \STATE Choose a nest among $n$ (say, $j$) randomly;
   \IF{$F_i < F_j$}
   \STATE Replace $j$ by the new solution;
  \ENDIF
  \STATE
   \STATE A fraction ($p_a$) of the worse nests are abandoned and new ones are built;
   \STATE Keep the best solutions/nests;
   \STATE Rank the solutions/nests and find the current best;
   \STATE Pass the current best solutions to the next generation;
\ENDWHILE

\end{algorithmic}

Levy flight is an essential part of the CS algorithm. It is defined by the following equation

\begin{equation}
\label{Levi1}
X_i(t+1) = X_i(t) + \alpha \bigotimes Levy(\lambda),
\end{equation}

where $\alpha~(\alpha>0)$ represents a step size. Eq. \ref{Levi1} basically represents a stochastic equation for a random walk which is a Markov chain. More precisely, the next position or function value that shall be evaluated depends on two parameters; namely, the current position and transition function ($\alpha \bigotimes Levy(\lambda)$). The next position (status) depends only on the current position ($X_i(t)$) and probability of transition ($\alpha \bigotimes Levy(\lambda)$). Levy($\lambda$), which is the random step length, is drawn from a Levy distribution. The consecutive positions generated through steps of the Levy flight create a random walk that has a heavy  tail distribution. The use of Levy flight instead of a simple random walk significantly improves the performance of CS.

As presented in this pseudo code, the CS algorithm has several stages. The first stage is to generate the initial population. In the main loop of the algorithm a new solution $F_i$ is generated using Levy flight from a random nest $i$. If this solution is better than a one belonging to random nest $F_j$, the $F_j$ substituted by $F_i$. Finally, in the goal of increasing the diversity of the search, a fraction $p_a$ of the worst nests is changed using Levy flight with a larger step applied to avoid trapping in local optimal solutions.

The main advantage of the CS search is its robustness and its dependence on only a few number of parameters that are needed to be fine tuned. Certainly, these parameters depend on the function that is being optimized like in the case of PSO and GA.

\subsection{Nelder-Mead method}

The Nelder–Mead simplex method, frequently called the downhill simplex method or amoeba method, is widely used for finding local minima solutions, for well defined problems for which a derivative is not known. The NMS falls in the general class of direct search methods.The basic component of the algorithm is a simplex. In case of a two parameter function, a simplex is a triangle, and the method results in a pattern search that is dependent of the values of the three vertices of the triangle. In a similar way for a $N$ dimension function a simplex will have $N+1$ vertices. In general, the method replaces some of the worst points (which have the largest values of the $f$) of the simplex with a new vertices that are acquired using a heuristic approach. In this way the simplex is transformed to a new one which is potentially closer to the minima. There are several potential transformation refection, expansion, contraction and shrinkage that define the generation of the search \cite{Nelder01011965}. In NMS algorithm, the heuristic choice and the application of the transformation are called a flip.
\begin{figure}
\centering
\includegraphics[width=.5\textwidth]{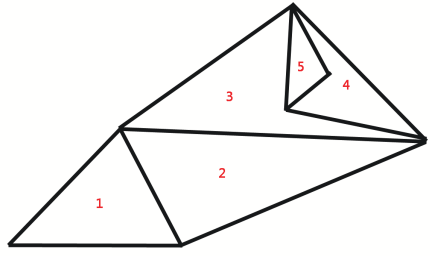}
\caption{Illustration of the movement of a 2D simplex using the flip operation inside of the NMS algorithm.}
\label{AL}
\end{figure}

More precisely the NMS algorithm is defined in the following way.  For function $f$ with  $n$ dimensions, a simplex will be a convex hull consisting of $n+1$ vertices. We will define a simplex  with vertices $x_1, x_2, .., x_n, x_{n+1}$ by $\triangle$. The NMS generates iteratively a sequence of simplices by performing the following procedure. First all of the vertices inside of simplex $\triangle$ are sorted depending on the objective function value

\begin{equation}
		f(x_1)\leq f(x_2)\leq .. \leq f(x_n)
\end{equation} 

At each of iteration some of the vertices are removed from the simplex and new ones are added. Using the following steps 

\begin{itemize}

\item Reflection

The reflection point $x_r$ is calculated using the following formula

\begin{equation}
       \label{Reflection}
		x_r = x_c + \alpha( x_c - x_{n+1}) 
\end{equation} 

In Eq. \ref{Reflection} $x_c$ represents the centroid of the simplex calculated in the following way

\begin{equation}
		x_c = \frac{1}{n+1}\sum_{i=1}^{n+1} x_i
\end{equation} 

In case the value of $f_r = f(x_r)$ satisfies the condition $f_1\leq f_r\leq f_n$ we replace $x_{n+1}$ with $x_r$ and go to next iteration. 

\item Expansion

In case $f_r < f_i$, an additional step is done to attempt by calculating an expansion vertex $x_e$.

\begin{equation}
       \label{Extenstion}
		x_e = x_c + \beta(x_r - x_c) 
\end{equation} 

If $f_e = f(x_e)$ satisfies $f_e < f_r$ replace $x_{n+1}$ with $x_e$ otherwise with $x_r$.
\item Outside Contraction 

In case $f_n \leq f_r < f_{n+1}$, compute the outside contraction point.
\begin{equation}
       \label{OutC}
		x_{oc} = x_c + \gamma(x_r- x_c) 
\end{equation} 

If $f_{oc} = f(x_e)$ satisfies $f_{oc} \leq f_r$ replace $x_{n+1}$ with $x_{oc}$ otherwise do a shrink operation.

\item Inside Contraction

In case  $ f_r \geq f_{n+1}$, compute the inside contraction point.

\begin{equation}
       \label{InC}
		x_{ic} = x_c - \gamma(x_r- x_c) 
\end{equation} 

If $f_{ic} = f(x_{ic})$ satisfies $f_{ic} \leq f_{n+1}$ replace $x_{n+1}$ with $x_{ic}$ otherwise do a shrink operation.

\item Shrink

\begin{equation}
        \label{Shrink}
		x_i = x_1 + \delta(x_i- x_1) 
\end{equation} 
\end{itemize}

To fully define the  flip operation it is necessary to specify the values of $\alpha$, $\beta$, $\gamma$ and $\delta$.  Their values  should satisfy the following conditions.
\begin{equation}
\alpha>0,\,\,\,\, 0<\beta<1, \,\,\,\, \gamma>1, \,\,\,\, \gamma>\alpha, \,\,\,\, 0< \delta <1. 
\end{equation} 

In the standard implementation these parameters have the following values: 
$\alpha=1$, $\beta=0.5$, $\gamma=2$ and $\delta=0.5$. In our hybridization of NMS we have used these values. We wish to mention that the NMS algorithm, and consequently the proposed hybridization, can be further improved by incorporating adaptive parameters \cite{gao2012implementing}.

\subsection{Hybridized Methods}

There are two main disadvantages of the CS algorithm. First, it does not incorporate any type of local search to increase the convergence speed when it is close to some local or global minima,  but only uses the Levy flight to generate new test positions. Contrary to this,  GA and PSO are much faster in narrowing to optima when they are close. The second problem is that the performance of CS is highly dependant on the value of $\alpha$ in Eq. \ref{Levi1}. It has been shown that its value does not only depend to the on scale the test function, but for high precision of results, its value needs to adapt during the execution of the algorithm.

There has been a wide range of approaches to improve the performance of CS by hybridization. It has been shown that combining CS with other types of population based algorithms is a good approach. Some of the most successful ones are its combination with GA   \cite{zheng2013cooperative}, PSO \cite{ghodrati2012hybrid}, differential evolution \cite{wang2012hybrid}, and ant colony optimization \cite{CSACO}. Although this type of hybridization manages to improve the performance of the original method, the implementation becomes significantly more complex than the basic CS. On the other hand, it could be argued that the new methods, which couples two population based techniques, that are generally more suitable for globalized searches, combines to very similar approaches. It is also important to mention that very limited research has been done in the hybridization of CS with some local search methods, which has proven to be very profitable for other population based algorithms.

In our work, we attempt to improve the performance of CS, by hybridization with NMS algorithm while maintaining the simplicity and robustness of the original algorithm. Previously, population based algorithms have been successfully combined with NMS \cite{Zahara20093880, Fan2006401, KoduruDW06}. There are two main directions to this type of hybridization; namely, cascaded or tandem \cite{KoduruDW06}. In the cascaded approach, the two methods are used consecutively, and resulting in a more globalized search. In case of a tandem hybridization, the two methods influence each others performance and generally result in a more localized search which is more suitable for functions with a lower number of parameters. In the case of optimization of solar cells, the functions of interest would have from 3-10 parameters, so we have focused on developing a more localized search using the tandem approach. The focus of the method is to find solutions of good quality with a low number of functions evaluations. The incentive for this is the fact that the calculation of efficiency of multi layer solar cells is computationally very expensive.

\section{Hybridized method}

As previously mentioned, the goal is to find the minima of a function containing a relatively small number of parameters (3-10). To achieve this, we have hybridized the NMS using Cuckoo search. The general idea is to exploit the effectiveness of NMS as a local search algorithm, but combine it with the CS to get a wider search area. By combining CS with NMS, the original method becomes more capable in narrowing down to good solutions. One of the problems of CS is that except for the focusing of the search by  the replacement of lower quality solutions with better ones, the search is relatively random. In the new approach, we wish to search the solution space in a more structured way. This is done by incorporating the flip operation from NMS to CS.

The changes to the original CS method are the following:

\begin{itemize}

\item{Nests in the new method are simplexes instead of parameter values that correspond to good solution. As a consequence the initial population will be a set of simplexes.}
\item{While in the original CS, Levy flight is used to explore new solutions, in the new method, this is done using the flip operation of NMS. More precisely, at each step of the algorithm, nest $i$ is transformed to nest $i'$ by using a flip.}
\item{In the presented approach, only a part of the vertices inside of a simplex are replaced. This is because a simple replacement of lower quality nests with better ones in the case of using simplexes combined with the flip operation is not productive. This is due to the fact that NMS has a deterministic search pattern, and if this was done, we would be just repeating the tests for same solutions.}
\end{itemize}

The new approach can be formalized by the following pseudo code.

\begin{algorithmic}
\STATE
\STATE{Objective function: $f(X)$, $X=(x_1, x_2,.., x_d)$}
\STATE
\STATE Generate an initial population of  $n$ simplexes;
\STATE Test $dn$ solutions;
\WHILE{($t<MaxGeneration$) or ($stop criterion$)}
   \STATE Get a cuckoo randomly (say, $i$) and replace its using a NMS flip;
   \STATE Evaluate its quality/fitness $F_i$
    \STATE
   \STATE Choose a nest among $n$ (say, $j$) randomly;
   \IF{$F_i < F_j$}
   \STATE Replace worst $p$ vertices in simplex $j$
   \STATE  by the vertices from simplex $i$;
  \ENDIF
  \STATE
   \IF{$mod(t,k)=1$}
   \STATE A fraction ($p_a$) of the worse simplex are abandoned and new ones are built;
   \ENDIF
   \STATE Keep the best solutions/nests;
   \STATE Rank the solutions/nests and find the current best;

\ENDWHILE
\end{algorithmic}

The initial simplexes are generated around randomly selected points, with a relatively small sizes. We use a slight modification of the approach for generating simplexes proposed by L. Pfeffer, which is also used in MatLab implementation of NMS \cite{InitSimplex}. A simplex  $\triangle$ with vertices $x_i$ is generated around a vertex $a = (a_1, a_2, .., a_n)$. All of the coordinates of $x_i$ are equal to the ones of $a$ except the one on position $i$, which is calculated as

\begin{equation}
  \label{RandomSimplex}
  x = a_i+ \alpha a_i
\end{equation}

In Eq. \ref{RandomSimplex} $x$ represents the value of $x_i$ at position $i$. $\alpha$ is a random parameter that is used generate simplexes of different sizes. In our test the best results where achieved when $\alpha$ had a uniform distribution inside of the  interval $(0,0.25)$

As it has been previously mentioned, the flip operation is simply used instead of the Levy flight, as it can be seen in the pseudo code. The substitution of nest of lower quality is done by coping some of the vertices of the lowest quality by some of the high quality ones from the better simplex.

At this stage, there are two important details that must be pointed out. First, the best vertex from the higher quality nest (simplex) is not used because this would make the search  focus very quickly. If the other high quality vertexes are used, the search focuses towards good solutions but still remains diverse. This stage has two possibilities, either a relatively random search of the space in the direction of the higher quality nest if the simplexes are distant or a wider search around the good solution if the simplexes are close to each other. The algorithm has two parameters $k$ and $p$ that are used for fine tuning of the method. The part of vertexes $p$ that should be overwritten is dependent on the number of parameters of the function that we wish to minimize. In our experiments, it has been shown that for functions with a low number of parameters, the value $p=1$ vertices is a good choice. In general, if this number is too low the search needs many steps to actually move to the region of good solutions, and contrary if it is two large this movement is too fast. The second tuning parameter $k$ is used to specify how randomized the search will be, or in other words, how frequently new random nests will be generated. In our test it has been shown that having $k$ equal to two times the number of nests is a good choice.

\section {Experiments and Results}

In this section, we present results of using the new method to optimize two sets of problems. The first test are conducted on standard benchmark test functions and compared to previously published results \cite{Fan2006401}. More precisely, we have compared our method to the hybridizations of NMS with PSO and NMS with GA on problems of 3-10 parameters. Since PSO and GA are methods that often competing methods CS, this gives us a good evaluation of the new hybridized method.

To fully evaluate the effect of the proposed hybridization of the CS, in the second section we present a comparison to other version of the CS.
In the final part of our tests, we analyze the performance of the new method on a practical problem of optimization of multijunction and split spectrum solar cells. We have tested the effectiveness of our method for systems with a 3-10 different cells of both types. To be able to have a quantitative evaluation of the new method, it has been compared to NMS, simulated annealing and GA.

The method has been implemented by creating code using MatLab R2013a. The calculations have been done on a machine with Intel(R) Core(TM) i7-2630 QM CPU \@ 2.00 GHz, 4GB of DDR3-1333 RAM, running on Microsoft Windows 7 Home Premium 64-bit.

\subsection{Comparison to  other Hybridizations of NMS}

In our first group of test, we compare the performance of the new method to hybridization of NMS with two other population based meta heuristics, GA and PSO. Test has been done on 10 different test functions as presented by Fan et al. \cite{Fan2006401} and Chelouah and Siarry \cite{Chelouah2003335} where their definitions can be found. Tests for functions with two variables have been done for the Branin RCOC ($RC$), B2, Goldstein and Price ($GP$), Shubert ($SH$),  Zakharov ($Z_n$); with three variables Hartmann ($H_{3,4}$), and with four variables on the  Shekel $S_{4,3}$. The test has also been done on the  Rosenbrock ($R_n$) function for 2, 5 and 10 parameters. For each of the test functions, 100 runs of the NMS-CS algorithm have been done with different initial simplex positions similarly to what is done in article \cite{Fan2006401}. In the two competing algorithms the initial vertices are randomly selected inside of the specified  search domain, while in the case of the proposed hybridization these vertices are used for generating the initial simplixes using the method presented in the previous section.     In the results section of the mentioned article, "successful" runs are  defined for solutions that have a certain level of precision. We have excluded this measure from the results given in Table \ref{table:NecReloc}, since our method has had a 100 successless runs for each of the 10 test functions; as it had been the case with  NMS-GA and NMS-PSO given in \cite{Fan2006401}. In our test, we have used the same stopping criteria as in article \cite{Fan2006401} or when a high enough precision is reached. The stopping criterion is given in the following equation

\begin{equation}
	S_f = \sqrt{\sum_{i=1}^N f(x_i) - \overline{f}}< \varepsilon
\end{equation}

Where $\overline{f}  = \frac{1}{N+1}\sum_{i=1}^N f(x_i)$ is the average of the best $N$ solutions. This criterion is  based on the standard deviation of the objective function values over $N$ best solutions.The chosen value for $\epsilon = 10^{-7}$. In the case of the previous published work $N$ was equivalent to one third of the population. In the case of the proposed method $f_(x_i)$ would represent the best solution inside of simpex $i$, and consequently we would only observe the best third ($N$) simplexes.

The NMS-CS has used 6 nests for all the test functions except $R_{10}$ where it was 20. This is lower then the recommended 25 nests for the CS which is used in a wide range of articles. We believe that NMS-CS has achieved the best results when using a much smaller number of nest due to fact that it uses the NMS flip operation to narrow in on solutions. We can see the comparison of the three methods in Table \ref{table:NecReloc}

\begin{table}[htb]
\footnotesize
\center
\caption{Comparison of population based hybridizations of the NMS algorithm on standard benchmark functions. The best results have been underlined.}
\label{table:NecReloc}
\begin{tabularx}{350pt }{X X X X X X X}
\toprule
Test Function &\multicolumn{2}{c}{NMS-GA}&	\multicolumn{2}{c}{NMS-PSO}	&\multicolumn{2}{c}{NMS-CS}\\
 &Num. Eval.&Avg. Error&Num. Eval.&Avg. Error	 &Num. Eval.&Avg. Error\\

\midrule
RC	&356&4.0e-5&	230&1.0e-4&269 &2.1e-5\\
B2	&529&4.0e-5&	325&1.0e-5$>$&	\underline{132}&	\underline{1.0e-5$>$}\\
GP	&422&2.0e-5&	304&3.0e-5&313&2.4e-5	\\
SH	&1009&2.0e-5&	753&3.0e-5&\underline{569}&	\underline{2.0e-5}\\
$R_2$	&738&6.0e-5&	440&5.0e-5&473&	2.1e-5\\
$Z_2$	&339&4.0e-5&	186&1.0e-5$>$&	\underline{150}&	\underline{1.0e-5$>$}\\
$H_{3,4}$	&688&5.0e-5&	436&1.2e-4&	418&5.0e-4	\\
$S_{4,5}$	&2366&1.6e-4&	850&6.0e-5&	1125&	2.8e-5\\
$R_{5}$ &	3126&9.0e-5&	2313&4.0e-5&\underline{1504}&	\underline{3.1e-5}\\
$R_{10}$	&5194&2.0e-4&	3303&1.2e-4&2621	&	2.2e-4\\
\bottomrule
\end{tabularx}
\end{table}

It is observed that the new method manages to outperform NMS-GA. This confirms the conclusion form article \cite{Fan2006401}, that combining NMS with swarm intelligence methods is more beneficial than with genetic algorithms.  It is apparent that the new method has a very similar performance to NMS-PSO when the number of iterations is considered, but overall the new method manages to get slightly higher quality of results. More precisely the new method has managed to get clearly better results, in both the number of iterations and achieved precision, in four out of ten cases.      Due to the definition of benchmarks tests that have been previously published, to which we have made a comparison to, it was not possible to  strictly conclude which method was the best preforming in the rest of the cases but it can be said that the performance of NMS-PSO and NMS-CS was very similar.   It is important to point out, that the new approach is much simper to implement compared to the other two hybridizations of NMS.

\subsection{Comparison to other versions of Cuckoo Search Algorithms}

To  evaluate the performance of the proposed hybridization of the CS,  a comparison to other versions of the CS has also been conducted. More precisely the new method has been compared to the original CS algorithm proposed by Yang (Yang) \cite{yang2009cuckoo} and the modified version of the algorithm proposed by Walton (MCS) \cite{Walton2011710}. The comparison has been conducted on the benchmark functions used at the CEC 2013 \cite{liang2013problem}. The benchmark instances are divided into 3 groups, functions 1-5 are unimodal, 6-20 are multimodal and functions 21-28 are composite functions generated using the three to five of the functions of the two first groups.  The three methods have been compared for test functions having 2, 5 and 10 dimensions. The numerical experiments have been done using the code made available by the authors of the two competing methods \cite{CodeYang, CodeWolton}.

For each of the algorithms  50 independent runs have been performed for each of the test functions. We observed the average and  best found solution find for the 50 runs, with a fixed number of fitness function evaluations. In case of 2, 5 and 10 dimensional functions, a maximum of 2000, 5000 and 20000 function evaluations was allowed. To better evaluate the performance of the methods the best, worst, average  and standard deviation of the generated solutions is presented.  To have the a fair comparison we have done extensive fine tuning of the parameters that specify the two algorithms for problems of these sizes through a large number of tests, of course this was also been done for the proposed method.  The number of nests used for the Yang's and the new algorithm  was 6, 10 and 25 for test functions of 2, 5 and 10 dimensions. In case of MCS, doubling the number of the nest for each of the dimension has shown to give the best results. This is due to the fact that MCS only uses a high number of nest in the early steps of the algorithm. To evaluate the effectiveness we The results of these experiments are given in Tables \ref{table:Dim2}, \ref{table:Dim5}, \ref{table:Dim10}.

\begin{landscape}

\begin{table}[htb]

\footnotesize
\center
\caption{Comparison of different CS algorithms for benchmark functions with two dimensions. The best results have been underlined. }
\label{table:Dim2}
\begin{tabularx}{580pt }{X X X X X X X X X X X X X X X X}
\toprule

   &\multicolumn{4}{c}{Yang}&	\multicolumn{4}{c}{MCS}	&\multicolumn{4}{c}{NMS-CS}\\
   \cmidrule{3-4}     \cmidrule{7-8}  \cmidrule{11-12}\\
   & Min& Max&Avg &Std & Min& Max&Avg&Std& Min& Max&Avg&Std\\ 
\midrule
f1 &
\underline{0.0e+0} &\underline{0.0e+0} &\underline{0.0e+0} &  0.0e+0 & 3.6e-10 &2.1e-4 &2.5e-5 &  4.5e-5 & \underline{0.0e+0} &\underline{0.0e+0} &\underline{0.0e+0} &  0.0e+0\\
f2 &
2.9e-7 &1.7e+0 &8.5e-2 &  3.4e-1 & 3.2e+0 &1.1e+4 &1.5e+3 &  2.2e+3 & \underline{0.0e+0} &\underline{0.0e+0} &\underline{0.0e+0} &  0.0e+0\\
f3 &
1.4e-6 &8.1e+3 &1.6e+2 &  1.1e+3 & 6.5e+0 &5.7e+3 &1.1e+3 &  1.2e+3 & \underline{0.0e+0} &\underline{2.3e-13} &\underline{0.0e+0} &  3.2e-14\\
f4 &
6.0e-8 &1.9e-2 &1.6e-3 &  3.5e-3 & 1.1e-1 &1.5e+4 &2.0e+3 &  3.3e+3 & \underline{0.0e+0} &\underline{0.0e+0} &\underline{0.0e+0} &  0.0e+0\\
f5 &
\underline{0.0e+0} &\underline{0.0e+0} &\underline{0.0e+0} &  0.0e+0 & 3.7e-7 &3.7e-3 &4.9e-4 &  7.1e-4 & \underline{0.0e+0} &\underline{1.1e-13} &\underline{0.0e+0} &  3.7e-14\\
f6 &
8.0e-13 &7.5e-2 &2.1e-3 &  1.1e-2 & 6.5e-9 &1.3e-1 &1.1e-2 &  2.4e-2 & \underline{0.0e+0} &\underline{0.0e+0} &\underline{0.0e+0} &  0.0e+0\\
f7 &
8.6e-7 &\underline{3.2e-2} &\underline{3.0e-3} &  6.9e-3 & 1.0e-3 &1.8e+0 &2.2e-1 &  4.8e-1 & \underline{8.1e-10} &3.6e-1 &2.0e-2 &  6.4e-2\\
f8 &
2.1e-6 &2.0e+1 &5.4e+0 &  7.4e+0 & 1.0e-3 &1.8e+1 &1.0e+0 &  3.4e+0 & \underline{8.0e-13} &\underline{2.0e+1} &\underline{7.2e-1} &  3.4e+0\\
f9 &
1.3e-4 &9.1e-1 &1.6e-1 &  2.5e-1 & 7.5e-4 &5.3e-1 &8.8e-2 &  1.4e-1 & \underline{0.0e+0} &\underline{7.9e-1} &\underline{4.9e-2} &  1.9e-1\\
f10 &
9.6e-5 &1.0e-1 &2.7e-2 &  2.1e-2 & 7.4e-3 &2.0e+0 &4.3e-1 &  5.1e-1 & \underline{0.0e+0} &\underline{1.2e-1} &\underline{2.6e-2} &  2.8e-2\\
f11 &
\underline{0.0e+0} &\underline{4.0e+0} &1.8e-1 &  6.3e-1 & 3.5e-9 &1.0e+0 &\underline{1.5e-1} &  3.6e-1 & \underline{0.0e+0} &\underline{9.9e-1} &1.8e-1 &  3.9e-1\\
f12 &
5.7e-14 &2.0e+0 &\underline{2.6e-1} &  4.7e-1 & 1.7e-6 &2.0e+0 &9.0e-1 &  4.6e-1 & \underline{0.0e+0} &\underline{1.0e+0} &2.9e-1 &  4.5e-1\\
f13 &
\underline{0.0e+0} &3.5e+0 &\underline{3.7e-1} &  8.8e-1 & 2.5e-7 &3.5e+0 &1.7e+0 &  1.2e+0 & \underline{0.0e+0} &\underline{2.8e+0} &5.4e-1 &  9.6e-1\\
f14 &
2.4e-8 &\underline{1.7e+1} &\underline{1.1e+0} &  3.3e+0 & 5.3e-7 &1.7e+1 &1.4e+0 &  4.0e+0 & \underline{0.0e+0} &1.8e+2 &1.6e+1 &  3.3e+1\\
f15 &
7.3e-5 &1.2e+2 &1.8e+1 &  3.1e+1 & 9.9e-5 &\underline{1.2e+2} &\underline{1.2e+1} &  2.3e+1 & \underline{0.0e+0} &1.2e+2 &1.5e+1 &  2.6e+1\\
f16 &
1.2e-1 &2.3e+0 &9.5e-1 &  5.7e-1 & 1.3e-1 &1.9e+0 &8.2e-1 &  4.3e-1 & \underline{0.0e+0} &\underline{2.7e-1} &\underline{6.2e-3} &  3.8e-2\\
f17 &
2.4e-5 &\underline{2.5e+0} &\underline{1.1e+0} &  8.2e-1 & 4.9e-5 &3.2e+0 &1.3e+0 &  1.0e+0 & \underline{0.0e+0} &3.2e+0 &1.6e+0 &  9.7e-1\\
f18 &
1.4e-1 &3.1e+0 &1.6e+0 &  8.2e-1 & 1.3e-4 &\underline{5.2e+0} &\underline{1.4e+0} &  1.1e+0 & \underline{0.0e+0} &3.1e+0 &1.7e+0 &  9.4e-1\\
f19 &
\underline{0.0e+0} &\underline{4.9e-2} &\underline{8.4e-3} &  1.4e-2 & 1.3e-9 &1.8e-1 &4.7e-2 &  4.4e-2 & \underline{0.0e+0} &1.2e-1 &1.3e-2 &  2.0e-2\\
f20 &
4.3e-6 &8.6e-1 &3.5e-2 &  1.2e-1 & 1.9e-2 &1.6e-1 &4.0e-2 &  3.4e-2 & \underline{0.0e+0} &\underline{2.5e-1} &\underline{3.4e-2} &  4.6e-2\\
f21 &
1.3e-12 &2.0e+2 &3.8e+1 &  7.3e+1 & 7.0e-4 &1.0e+2 &\underline{4.3e+0} &  2.0e+1 & \underline{0.0e+0} &\underline{1.0e+2} &5.8e+0 &  2.3e+1\\
f22 &
2.9e-11 &\underline{1.6e+2} &\underline{1.2e+1} &  3.0e+1 & 6.6e-5 &1.2e+2 &1.8e+1 &  3.4e+1 & \underline{1.1e-13} &1.6e+2 &4.7e+1 &  4.5e+1\\
f23 &
1.6e-8 &2.9e+2 &7.9e+1 &  9.8e+1 & 1.8e-3 &\underline{1.0e+2} &\underline{2.0e+1} &  2.8e+1 & \underline{0.0e+0} &2.8e+2 &6.2e+1 &  6.7e+1\\
f24 &
1.3e-8 &\underline{1.0e+2} &1.0e+1 &  2.3e+1 & 4.0e-4 &2.4e+1 &8.2e+0 &  9.0e+0 & \underline{0.0e+0} &7.2e+1 &\underline{7.9e+0} &  1.5e+1\\
f25 &
2.7e-9 &1.1e+2 &4.5e+1 &  4.9e+1 & 1.7e-3 &1.1e+2 &\underline{3.2e+1} &  4.3e+1 & \underline{0.0e+0} &\underline{1.0e+2} &4.0e+1 &  4.8e+1\\
f26 &
1.6e-9 &\underline{1.0e+2} &5.6e+0 &  2.0e+1 & 4.8e-6 &3.2e+1 &1.7e+0 &  4.8e+0 & \underline{0.0e+0} &2.3e+1 &\underline{1.3e+0} &  3.4e+0\\
f27 &
3.6e+0 &3.1e+2 &1.5e+2 &  8.9e+1 & \underline{7.9e-2} &1.4e+2 &8.9e+1 &  4.8e+1 & 8.5e-1 &\underline{1.5e+2} &\underline{7.9e+1} &  4.9e+1\\
f28 &
2.1e-10 &3.0e+2 &4.0e+1 &  9.3e+1 & 1.0e-2 &1.0e+2 &\underline{2.7e+1} &  4.3e+1 & \underline{0.0e+0} &\underline{1.0e+2} &3.2e+1 &  4.7e+1\\ 

\bottomrule
\end{tabularx}
\end{table}
\end{landscape}

\begin{landscape}

\begin{table}[htb]
\footnotesize
\center
\caption{Comparison of different CS algorithms for benchmark functions with five dimensions. The best results have been underlined. }
\label{table:Dim5}
\begin{tabularx}{580pt }{X X X X X X X X X X X X X X X X}
\toprule

   &\multicolumn{4}{c}{Yang}&	\multicolumn{4}{c}{MCS}	&\multicolumn{4}{c}{NMS-CS}\\
   \cmidrule{3-4}     \cmidrule{7-8}  \cmidrule{11-12}\\
   & Min& Max&Avg &Std & Min& Max&Avg&Std& Min& Max&Avg&Std\\ 
\midrule
f1 &
\underline{0.0e+0} &\underline{0.0e+0} &\underline{0.0e+0} &  0.0e+0 & 5.3e-6 &1.2e-1 &3.9e-3 &  1.7e-2 & \underline{0.0e+0} &\underline{0.0e+0} &\underline{0.0e+0} &  0.0e+0\\
f2 &
1.0e+1 &1.1e+4 &6.9e+2 &  1.6e+3 & 5.5e+2 &6.3e+5 &1.4e+5 &  1.6e+5 & \underline{9.1e-10} &\underline{9.5e-3} &\underline{2.8e-4} &  1.4e-3\\
f3 &
1.2e+0 &3.4e+6 &1.3e+5 &  5.4e+5 & 1.9e+2 &7.3e+7 &3.5e+6 &  1.2e+7 & \underline{5.2e-6} &\underline{1.5e+5} &\underline{4.0e+3} &  2.2e+4\\
f4 &
1.7e+0 &8.2e+2 &8.8e+1 &  1.5e+2 & 3.0e+2 &9.2e+3 &3.8e+3 &  2.4e+3 & \underline{1.1e-10} &\underline{5.8e-2} &\underline{1.3e-3} &  8.2e-3\\
f5 &
\underline{1.1e-13} &\underline{9.8e-11} &\underline{9.3e-12} &  1.6e-11 & 1.1e-4 &4.7e-2 &7.0e-3 &  9.5e-3 & 4.9e-9 &2.2e-5 &2.6e-6 &  5.1e-6\\
f6 &
2.7e-5 &2.7e+0 &\underline{5.6e-1} &  6.6e-1 & 1.1e-3 &5.1e+0 &1.7e+0 &  1.9e+0 & \underline{1.1e-13} &\underline{3.9e+0} &1.2e+0 &  1.8e+0\\
f7 &
\underline{2.6e-2} &\underline{3.2e+0} &\underline{6.7e-1} &  7.0e-1 & 4.6e-1 &5.5e+1 &1.1e+1 &  1.2e+1 & 4.7e-2 &2.3e+1 &4.6e+0 &  6.2e+0\\
f8 &
4.6e+0 &2.0e+1 &2.0e+1 &  2.2e+0 & 2.6e+0 &\underline{2.0e+1} &1.8e+1 &  5.3e+0 & \underline{1.5e-1} &\underline{1.4e+1} &1.9e+1 &  4.8e+0\\
f9 &
2.3e-1 &2.8e+0 &1.7e+0 &  5.8e-1 & 4.6e-1 &\underline{3.2e+0} &1.7e+0 &  6.3e-1 & \underline{8.0e-2} &3.1e+0 &\underline{1.6e+0} &  6.6e-1\\
f10 &
9.3e-2 &4.6e-1 &2.5e-1 &  9.0e-2 & 3.7e-1 &3.7e+1 &6.6e+0 &  7.3e+0 & \underline{1.4e-8} &\underline{2.4e-1} &\underline{8.7e-2} &  5.5e-2\\
f11 &
\underline{1.2e-5} &\underline{3.3e+0} &\underline{1.2e+0} &  9.3e-1 & 5.0e-4 &5.0e+0 &1.8e+0 &  1.0e+0 & 9.9e-1 &1.8e+1 &5.6e+0 &  3.7e+0\\
f12 &
1.2e+0 &1.1e+1 &5.3e+0 &  2.1e+0 & 1.5e-1 &2.3e+1 &7.9e+0 &  4.8e+0 & \underline{2.9e-11} &\underline{1.4e+1} &\underline{5.1e+0} &  3.7e+0\\
f13 &
1.1e+0 &\underline{1.5e+1} &\underline{7.0e+0} &  3.7e+0 & 1.0e+0 &2.8e+1 &1.4e+1 &  6.6e+0 & \underline{9.3e-10} &2.8e+1 &9.2e+0 &  4.8e+0\\
f14 &
1.2e+1 &3.2e+2 &1.2e+2 &  7.7e+1 & \underline{3.9e-1} &\underline{1.8e+2} &\underline{5.6e+1} &  6.1e+1 & 7.2e+0 &6.9e+2 &3.5e+2 &  1.6e+2\\
f15 &
6.7e+1 &6.0e+2 &3.5e+2 &  1.1e+2 & 7.1e+0 &\underline{5.1e+2} &\underline{1.9e+2} &  1.1e+2 & \underline{6.8e+0} &5.9e+2 &2.7e+2 &  1.6e+2\\
f16 &
6.5e-1 &2.0e+0 &1.2e+0 &  3.1e-1 & 3.8e-1 &1.9e+0 &1.0e+0 &  3.6e-1 & \underline{7.6e-4} &\underline{9.6e-1} &\underline{2.9e-1} &  2.1e-1\\
f17 &
2.9e+0 &1.4e+1 &9.5e+0 &  2.0e+0 & \underline{1.7e-1} &\underline{1.3e+1} &\underline{6.9e+0} &  1.9e+0 & 3.0e+0 &1.9e+1 &9.7e+0 &  3.1e+0\\
f18 &
3.4e+0 &1.9e+1 &1.3e+1 &  3.0e+0 & 4.3e+0 &1.8e+1 &1.2e+1 &  3.2e+0 & \underline{1.9e+0} &\underline{1.7e+1} &\underline{9.1e+0} &  2.8e+0\\
f19 &
1.4e-1 &9.0e-1 &4.3e-1 &  1.5e-1 & 2.1e-2 &2.1e+0 &5.2e-1 &  3.9e-1 & \underline{2.0e-2} &\underline{6.7e-1} &\underline{3.2e-1} &  1.6e-1\\
f20 &
2.7e-1 &1.5e+0 &1.1e+0 &  3.0e-1 & \underline{2.5e-1} &\underline{1.7e+0} &\underline{1.1e+0} &  3.4e-1 & 2.9e-1 &2.5e+0 &1.4e+0 &  6.2e-1\\
f21 &
1.0e+2 &\underline{3.0e+2} &\underline{1.9e+2} &  9.8e+1 & 1.0e+2 &3.1e+2 &2.7e+2 &  7.4e+1 & \underline{3.5e-5} &4.0e+2 &2.2e+2 &  1.1e+2\\
f22 &
1.6e+2 &5.4e+2 &3.3e+2 &  9.8e+1 & \underline{2.3e-1} &\underline{5.1e+2} &\underline{2.4e+2} &  1.3e+2 & 1.4e+2 &9.2e+2 &6.0e+2 &  1.8e+2\\
f23 &
1.4e+2 &8.4e+2 &5.7e+2 &  1.3e+2 & \underline{7.5e-1} &\underline{7.9e+2} &\underline{4.5e+2} &  1.7e+2 & 1.1e+2 &8.9e+2 &5.5e+2 &  1.9e+2\\
f24 &
4.8e+1 &2.1e+2 &1.3e+2 &  3.9e+1 & 9.1e+1 &1.4e+2 &1.2e+2 &  9.7e+0 & \underline{3.3e+1} &\underline{2.1e+2} &\underline{1.2e+2} &  3.0e+1\\
f25 &
\underline{1.0e+2} &\underline{2.1e+2} &\underline{1.2e+2} &  2.3e+1 & 1.1e+2 &1.4e+2 &1.2e+2 &  9.3e+0 & 1.0e+2 &2.0e+2 &1.2e+2 &  1.6e+1\\
f26 &
1.5e+1 &\underline{2.0e+2} &\underline{1.0e+2} &  2.4e+1 & 5.1e+1 &1.2e+2 &1.1e+2 &  1.3e+1 & \underline{9.8e+0} &2.0e+2 &1.0e+2 &  3.2e+1\\
f27 &
3.0e+2 &3.8e+2 &3.5e+2 &  2.0e+1 & 3.1e+2 &4.1e+2 &3.5e+2 &  2.4e+1 & \underline{3.0e+2} &\underline{4.0e+2} &\underline{3.4e+2} &  2.5e+1\\
f28 &
1.0e+2 &\underline{3.0e+2} &\underline{2.5e+2} &  8.4e+1 & 1.0e+2 &4.0e+2 &3.2e+2 &  6.4e+1 & \underline{3.8e+1} &4.0e+2 &2.8e+2 &  8.7e+1\\ 

\bottomrule
\end{tabularx}
\end{table}
\end{landscape}

\begin{landscape}

\begin{table}[htb]
\footnotesize
\center
\caption{Comparison of different CS algorithms for benchmark functions with ten dimensions. The best results have been underlined. }
\label{table:Dim10}
\begin{tabularx}{580pt }{X X X X X X X X X X X X X X X X}
\toprule

   &\multicolumn{4}{c}{Yang}&	\multicolumn{4}{c}{MCS}	&\multicolumn{4}{c}{NMS-CS}\\
   \cmidrule{3-4}     \cmidrule{7-8}  \cmidrule{11-12}\\
   & Min& Max&Avg &Std & Min& Max&Avg&Std& Min& Max&Avg&Std\\ 
\midrule

f1 &
1.6e-10 &6.5e-9 &1.3e-9 &  1.3e-9 & 2.4e-5 &3.4e-2 &2.1e-3 &  6.0e-3 & \underline{6.8e-13} &\underline{2.8e-9} &\underline{3.3e-10} &  7.0e-10\\
f2 &
1.4e+5 &1.6e+6 &5.4e+5 &  3.4e+5 & 5.8e+5 &7.3e+6 &3.4e+6 &  1.7e+6 & \underline{1.2e-2} &\underline{6.1e+3} &\underline{1.9e+2} &  8.6e+2\\
f3 &
7.7e+8 &4.1e+42 &8.1e+40 &  5.7e+41 & 4.0e+6 &1.8e+9 &2.6e+8 &  3.4e+8 & \underline{3.7e+0} &\underline{2.7e+6} &\underline{1.1e+5} &  4.3e+5\\
f4 &
2.4e+3 &2.7e+4 &8.3e+3 &  4.5e+3 & 4.9e+3 &1.7e+4 &9.8e+3 &  2.8e+3 & \underline{1.7e-5} &\underline{8.6e+0} &\underline{3.8e-1} &  1.5e+0\\
f5 &
\underline{5.5e-7} &\underline{1.6e-5} &\underline{3.2e-6} &  2.5e-6 & 5.1e-4 &2.2e-1 &1.4e-2 &  3.2e-2 & 5.2e-5 &3.4e-3 &5.9e-4 &  6.0e-4\\
f6 &
2.5e-1 &9.8e+0 &\underline{6.0e+0} &  4.2e+0 & 5.0e-2 &8.0e+1 &3.2e+1 &  3.2e+1 & \underline{5.1e-10} &\underline{9.8e+0} &6.4e+0 &  4.6e+0\\
f7 &
7.6e+0 &\underline{5.4e+1} &\underline{2.5e+1} &  9.9e+0 & 1.7e+1 &1.2e+2 &5.0e+1 &  2.2e+1 & \underline{5.0e+0} &8.2e+1 &3.4e+1 &  2.0e+1\\
f8 &
2.0e+1 &2.1e+1 &2.0e+1 &  8.6e-2 & \underline{2.0e+1} &2.1e+1 &\underline{2.0e+1} &  1.0e-1 & 2.0e+1 &\underline{2.1e+1} &2.0e+1 &  8.6e-2\\
f9 &
4.0e+0 &8.2e+0 &6.6e+0 &  1.0e+0 & 3.5e+0 &7.7e+0 &5.7e+0 &  1.1e+0 & \underline{2.7e+0} &\underline{9.8e+0} &\underline{5.2e+0} &  1.5e+0\\
f10 &
2.3e-1 &9.5e-1 &6.1e-1 &  1.4e-1 & 1.2e+0 &3.4e+1 &7.2e+0 &  6.0e+0 & \underline{1.2e-7} &\underline{1.6e-1} &\underline{3.1e-2} &  3.5e-2\\
f11 &
4.0e+0 &2.2e+1 &1.4e+1 &  4.1e+0 & \underline{1.4e-4} &\underline{2.9e+1} &\underline{3.5e+0} &  4.3e+0 & 4.0e+0 &3.8e+1 &1.6e+1 &  6.9e+0\\
f12 &
9.3e+0 &4.6e+1 &3.0e+1 &  7.4e+0 & 8.1e+0 &4.8e+1 &2.5e+1 &  1.1e+1 & \underline{3.0e+0} &\underline{3.2e+1} &\underline{1.5e+1} &  6.5e+0\\
f13 &
1.3e+1 &5.4e+1 &3.2e+1 &  9.2e+0 & \underline{1.1e+1} &7.4e+1 &3.6e+1 &  1.4e+1 & 1.4e+1 &\underline{5.4e+1} &\underline{3.1e+1} &  1.1e+1\\
f14 &
3.8e+2 &1.4e+3 &1.0e+3 &  2.5e+2 & \underline{7.0e+0} &\underline{5.5e+2} &\underline{2.2e+2} &  1.2e+2 & 3.6e+2 &1.5e+3 &1.0e+3 &  3.0e+2\\
f15 &
7.6e+2 &1.8e+3 &1.4e+3 &  2.2e+2 & \underline{2.0e+2} &\underline{1.3e+3} &\underline{7.1e+2} &  2.3e+2 & 2.2e+2 &1.7e+3 &9.0e+2 &  3.2e+2\\
f16 &
3.6e-1 &1.9e+0 &1.4e+0 &  3.2e-1 & 5.5e-1 &1.6e+0 &1.1e+0 &  2.4e-1 & \underline{1.4e-1} &\underline{1.3e+0} &\underline{5.7e-1} &  2.8e-1\\
f17 &
2.1e+1 &5.0e+1 &3.7e+1 &  6.9e+0 & \underline{4.5e+0} &\underline{2.9e+1} &\underline{1.6e+1} &  4.2e+0 & 1.2e+1 &2.9e+1 &1.9e+1 &  4.3e+0\\
f18 &
3.2e+1 &5.7e+1 &4.8e+1 &  5.9e+0 & 2.1e+1 &5.1e+1 &3.6e+1 &  7.3e+0 & \underline{1.3e+1} &\underline{4.1e+1} &\underline{2.1e+1} &  5.7e+0\\
f19 &
9.8e-1 &3.2e+0 &2.2e+0 &  4.7e-1 & \underline{3.0e-1} &2.3e+0 &1.1e+0 &  4.6e-1 & 3.2e-1 &\underline{1.8e+0} &\underline{8.8e-1} &  3.1e-1\\
f20 &
3.1e+0 &\underline{3.9e+0} &3.5e+0 &  2.2e-1 & \underline{2.2e+0} &4.2e+0 &\underline{3.5e+0} &  4.5e-1 & 2.6e+0 &4.5e+0 &3.6e+0 &  4.5e-1\\
f21 &
\underline{1.0e+2} &\underline{4.0e+2} &\underline{2.0e+2} &  1.0e+2 & 3.0e+2 &4.0e+2 &4.0e+2 &  1.4e+1 & 2.0e+2 &4.0e+2 &4.0e+2 &  2.8e+1\\
f22 &
3.4e+2 &1.5e+3 &1.0e+3 &  2.4e+2 & \underline{3.2e+1} &\underline{5.0e+2} &\underline{2.7e+2} &  1.5e+2 & 6.0e+2 &2.0e+3 &1.3e+3 &  3.0e+2\\
f23 &
7.1e+2 &1.9e+3 &1.5e+3 &  2.3e+2 & \underline{2.9e+2} &\underline{1.8e+3} &\underline{1.1e+3} &  2.9e+2 & 6.2e+2 &2.1e+3 &1.3e+3 &  2.9e+2\\
f24 &
1.5e+2 &\underline{2.2e+2} &\underline{1.9e+2} &  2.4e+1 & 1.2e+2 &2.3e+2 &2.0e+2 &  3.7e+1 & \underline{1.1e+2} &2.2e+2 &2.2e+2 &  1.5e+1\\
f25 &
1.7e+2 &2.2e+2 &2.2e+2 &  7.5e+0 & \underline{1.2e+2} &\underline{2.2e+2} &\underline{1.9e+2} &  3.4e+1 & 1.7e+2 &2.3e+2 &2.1e+2 &  8.2e+0\\
f26 &
1.3e+2 &2.0e+2 &1.6e+2 &  1.9e+1 & 1.1e+2 &\underline{2.0e+2} &\underline{1.4e+2} &  2.7e+1 & \underline{1.1e+2} &3.2e+2 &1.8e+2 &  5.6e+1\\
f27 &
4.3e+2 &5.7e+2 &5.3e+2 &  3.4e+1 & 3.4e+2 &5.7e+2 &4.4e+2 &  6.1e+1 & \underline{3.0e+2} &\underline{6.1e+2} &\underline{4.4e+2} &  8.8e+1\\
f28 &
1.0e+2 &3.0e+2 &\underline{2.6e+2} &  8.3e+1 & 1.0e+2 &8.2e+2 &5.7e+2 &  2.3e+2 & \underline{1.0e+2} &\underline{7.6e+2} &4.1e+2 &  1.8e+2\\ 
\bottomrule
\end{tabularx}
\end{table}
\end{landscape}

In these tables the value 0.0e+0 means that the maximal precision of MatLab has been reached. From observing the results in Tables \ref{table:Dim2}, \ref{table:Dim5} and \ref{table:Dim10} it is noticeable that the new method has the greatest advantage when compared to the two other methods in the case of uni-modal functions.   Generally the hybridization using NMS is most advantageous in  a general slope towards the global minima. These two positive effects correspond to the use of the incorporation of the NMS algorithm. The flip operation in a certain, loosely defined, sense approximates the gradient of the function. This flip operation gives two effects in the hybridized algorithm. In case of large simplexes, which are created  using the mixing of nests global tendencies of the function are included in the search. On the other had, in the case of small ones, the flip operation makes it possible to quickly converge to the nearby local or global optimal solution. 

The method has had a better performance in the case of functions of 2 and 5 dimensions. When it is compared to the original CS algorithm it presents a significant improvement. It has managed to get lower values for average and  minimal solutions, for many test functions, compared to Yang's algorithm. The proposed method has lowered the performance of the original algorithm in only a few problem instances. This is contrary to the MCS, which is  an overall improvement to the basic algorithm, but has for a wide range of test functions  achieved worse results than the original CS algorithm.   It is evident that the new method has a less consistent behavior that two competing versions of CS. This can be concluded when we observe the results presented  in the Tables \ref{table:Dim2}, \ref{table:Dim5} and \ref{table:Dim10}, where  the methods has a much higher number of best found  minimal solutions than average ones.

\subsection{Multi-cell solar system optimization}

\begin{figure}
\centering
\includegraphics[width=2.0in]{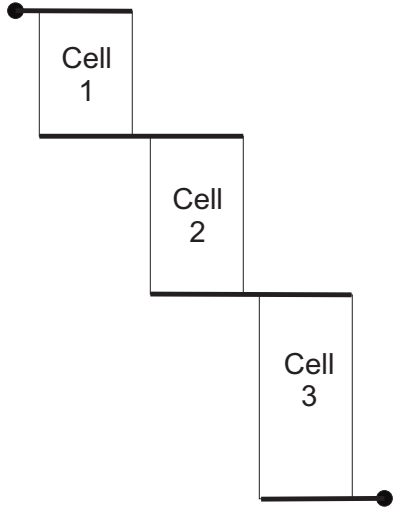}
\caption{The general concepts of multijunction solar cell systems where the dots are the collection (injection) points.}
\label{MJSC}
\end{figure}
\begin{figure}
\centering
\includegraphics[width=3.0in]{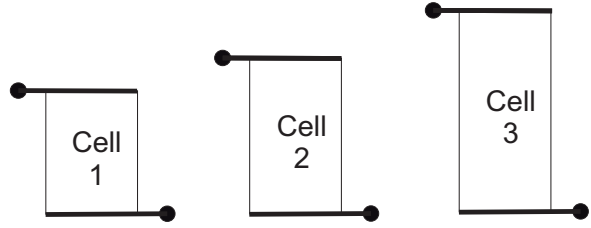}
\caption{The general concepts of split spectrum solar cell systems, where the dots are the collection (injection) points.}
\label{SSSC}
\end{figure}

In this section, we applied the presented method to the practical problem of optimizing the performance of multi-cell solar systems and mainly on multi-junction and  split spectrum solar cells. Finding the optimal efficiency of such cells is of high complexity due to the fact that the underlining equations are highly nonlinear. Previously,  research has been conducted in finding limits for different mathematical models.\cite{ComplexMultiBand1, ComplexMultiBand2}  These problem of finding the  In multi-junction solar cells, few cells of different energy gaps ($E_g$) are stacked in layers where in between buffer layers are used to allow transporting the photogenerated carriers between the cells. This type of cells is schematically represented in Figure-\ref{MJSC} and it is conceptually a two-terminal system with a series stack of two-level cells. As $E_g$s differ, each layer (or cell) abosrbs a portion of the incident solar radiation (i.e. for photonenergy above the layer $E_g$). Other photons are transimited to the next layer. This selective absortion/transmission is the mean for spectrum division.

Split spectrum solar cell system, follow a different approach. In this system, the idea is to split solar radiation by a pre-optical setup and then direct each of the split spectrum into a cell with matching $E_g$. We can see an illustration of this type of cells in Figure-\ref{SSSC}.

Practically, there are few differences between the two concepts. The main difference is current continuity. In split-spectrum system, the photogenerated carriers are collected separately for each cell. So, there is no physical constraint in this regard. However, in multi-junction cells, the photogenerated current from ny cell should be reinjected in the next one. Hence, it is essential to maintain the continuity.

There have been several strategies to estimate the practical efficiency limits of cells of this type.  Estimating the upper limit for multi-cell devices such as multi-junction and split spectrum, can be achieved by extending the single junction model  \cite{ADDED04,ADDED05,ADDED06,ADDED07}. In this work, we use the detailed balance model develop by Hanna and Nozik \cite{ADDED05,ADDED06}.

In both multi-junction and solar cell, the spectrum is ideally split based on the $E_g$s of the different cells. For $N$-cell system where the gaps are ordered ascendingly, the photogenerated generated current in the i$^{th}$ cell is

\begin{equation}
\label{Jgn}
	J_g^{(i)} = q \int_{E_g^{(i)}}^{E_g^{(i+1)}} \phi (E) dE
\end{equation}
$q$ is the electron charge, and $E$ is photon energy (in $eV$), and $\phi (E)$ is the standard AM1.5G flux. For the last cell, the upper limit is $\infty$. In the ideal case, it is assumed that radiative recombination is the only cause for recombination. So, the recombination current is calculated based on the generalized black body radiation as follow
\begin{equation}
\label{Jrn}
	J_r^{(i)} = q a \int_{E_g^{(i)}}^\infty  \frac{E^2}{\exp\left( \frac{E - \gamma(E) V^{(i)}}{k T} \right) - 1}  dE
\end{equation}
Hence, the resulted overall generated current in the i$^{th}$ cell is
\begin{equation}
\label{Jn}
	J^{(i)} \left( E_g, V, T \right) = J_g^{(i)} - J_r^{(i)}
\end{equation}
Then for each cell, the conversion efficiency becomes
\begin{equation}
\label{Effn}
	\eta^{(i)} \left( V^{(i)} \right) = \frac{V^{(i)} \, J^{(i)}}{P_{in}}
\end{equation}

In split spectrum cells, the photogenerated current is extracted separately for each cell. So, the total efficiency is
\begin{equation}
\label{Effss}
	\eta^{(SS)} = \sum_i \eta^{(i)} = \frac{1}{P_{in}} \sum_i V^{(i)} \, J^{(i)}
\end{equation}
For multi-jucntion, it is necessary to have the photogenerated current flow from one cell to the other before extracted in the external terminals. This series connection requires that the current must be the same in all of them and it is equal to the lowest current achieved in any of the cells. This fact resulted in changing the total efficiency to
\begin{equation}
\label{Effmj}
	\eta^{(MJ)} = \sum_i \eta^{(i)} = \frac{1}{P_{in}} J_{min} \sum_i V^{(i)}
\end{equation}
The goal of the optimization is to maximize $\eta$ and hence for practical application, we are actually minimizing the value $1-\eta^{(MJ)}$.

In the experiments, the performance of the new method is compared with NMS, SA and GA for split spectrum and multi-junction solar cells. The tests are conducted for solar cells that have 4 to 10 layers. For each of the problem sizes, we have compared the quality of results that can be archived using 1500 function evaluations. For NMS, SA and GA, methods that are a part of the MatLab toolbox are used. We have fine 

For the SA and the GA we have conducted a large number of tests on the smallest problem sizes to fine tune the parameters available in Matlab for defining these methods. When specifying the properties of the SA, we found that the optimal initial temperature for the optimization algorithm was 100.The temperature function used was set with the Matlab parameter TEMPERATUREEXP, in which the temperature at any given step was 0.95 times the temperature from the previous one. The reannaling interval was set to 100, the step has length square root of temperature, as defined by the Matlab parameter value ANNEALINGBOLTZ. In case of GA the population size was set to 15 out of which 2 where consider as elitist. The Crossover fraction was set to 0.8. The mutation was set by the Matlab parameter value MUTATIONUNIFORM, which correspond to selection rate of 0.01.

From physical properties of the problem, it is known that $E_g$ values should belong to the domain $(0,4)$. Solar radiation vanishes for photon energy above 4 $eV$. For the problem of interest, it is known that high quality solutions can be found close to $E_g$ values that equally divide the photogenerated carriers or if the corresponding voltages are equably divided in the expected range. In case of NMS, the search is started using 10 different starting points, out of which 2 correspond to the parameter values close to which it is expected to find good solutions and 8 random selection of parameters. In case of SA, only two starting points for which the highest quality of solutions is expected. We have used the same parameter sets as for NMS and 5 new random ones for the initial population. In case of NMS-CS, the same initial points as for GA have been used to generate the starting 15 nests (simplexes).

The calculation time for all of the methods was very similar due to the fact that the evaluation of the efficiency function is the most computationally expensive part. Because of the long execution time, only a single run has been conducted, in the sense that 1500 function evaluations have been done, for each of the methods. In table \ref{table:SSSC} and  \ref{table:MJSC} we can see the results acquired for split spectrum and multijunction solar cell respectably.

\begin{table}[htb]
\footnotesize
\center
\caption{\label{table:SSSC}Comparison of different meta heuristics for Split Spectrum Solar Cells. The best results have been underlined.}
\begin{tabularx}{350pt }{X X X X X X X}
\toprule
Number of cells &NMS&	SA	&GA&NMS CS\\
\midrule
3&51.348  &50.876&	\underline{51.351}&\underline{51.351}\\
4&55.387&53.562&	55.105&\underline{55.396}\\
5&57.633&55.419&	57.630&\underline{57.790}\\
6&59.110&58.594&	59.006&\underline{59.658}\\
7&60.290&59.553&	60.419&\underline{60.706}\\
8&61.247&60.228&	61.315&\underline{61.618}\\
9&61.871&61.075&	62.066&\underline{62.596}\\
10&62.350&61.517&	62.451&\underline{63.296}\\
\bottomrule
\end{tabularx}
\end{table}

In the case of calculating the efficiency of split spectrum solar sells, the use of SA has given results of lower quality than the other methods. NMS manages to get better results than GA in case of smaller problem instances, which can be explained by the fact that the smaller size of the solution space, make a local search method more effective. In all of the tested cases, NMS CS has managed to produce the results of the best quality.

\begin{table}[htb]
\footnotesize
\center
\caption{\label{table:MJSC}Comparison of different meta heuristics for multijunction solar cells. The best results have been underlined.}
\begin{tabularx}{350pt }{X X X X X X X}
\toprule
Number of cells &NMS&	SA	&GA&NMS CS\\
\midrule
3&50.961&50.548&50.719&\underline{51.003} \\
4&54.533&53.243&53.801&\underline{54.558} \\
5&56.183&55.086&52.141&\underline{56.610} \\
6&57.848&56.470&57.363&\underline{58.078} \\
7&59.714&58.556&59.423&\underline{59.732} \\
8&59.576&58.949&\underline{60.218}&60.140 \\
9&60.619&59.073&	\underline{60.896}&59.051\\
10&61.098&59.634&	\underline{61.461}&61.231\\
\bottomrule
\end{tabularx}
\end{table}

In case of multi junction solar cells, where the minimization problem becomes more complex due to the existences of constraints, similar results have been achieved. Although results are less conclusive, since only one test run has been performed, which can make the choice of the random seed influence the performance of the different methods. An example of this are the performance of GA and NMS CS in the case of 5 and 9 cells respectably, where the found efficiency is much lower than expected. We believe the reason for this is that the method has been trapped in a strong local minima. The new method has achieved better quality results in case of smaller problem instances, but has been slightly outperform by GA in case of the 3 problems of the greatest size. It is important to point out, that hybridized method has improved the performance of NMS in all but the one extreme case.

\section{Conclusion}

In this paper, we have presented a new method for the minimization of multi-parameter functions, with the focus on its application for the maximization of efficiency of multi-cell solar systems. The method is a hybridization of the CS with NMS algorithm. The new approach incorporates some fundamental parts of the NMS into CS. More precisely, it uses simplexes for nests in the CS, which makes it possible to use the NMS flip operation instead of the Levy flight. In this way, the good properties of NMS, as a local search, are preserved while the new methods manages to also conduct a effective exploration of a larger solution space. In comparison to the original CS algorithm, the hybridized approach becomes more capable for finding high quality solutions since it is combined to a local search method. A second advantage of the new method is the higher level of robustness, since it avoids using some tuning parameters that exist in CS.

The new approach has been tailored for the specific problem of solar cell optimization.  Although the function corresponding to this problem has a relatively low number of parameters 3-10, the complexity comes from the fact that the its evaluation is computationally expensive. In our experiments, we have shown that the new approach manages to outperform NMS, GA, and SA for this problem. To better evaluate the new method we have also compared its performance to other hybridizations of NMS using population based meta heuristics like GA and PSO. We have shown that the new method is very competitive, and in many cases outperforms, these algorithms on standard benchmark functions. One of the main advantages of the new method is its simplicity of implementation, especially if it is compared to other hybridizations of this type.


\end{document}